\documentclass[sn-mathphys]{sn-jnl}% Math and Physical Sciences Reference Style
\usepackage{color}
\jyear{2021}%

\theoremstyle{thmstyleone}%
\theoremstyle{thmstyletwo}%

\theoremstyle{thmstylethree}%

\begin{document}

\title[NL-CS Net: Deep Learning with non-local Pior for Image Compressive Sensing]{NL-CS Net: Deep Learning with Non-Local Pior for Image Compressive Sensing}

%%=============================================================%%
%% Prefix	-> \pfx{Dr}
%% GivenName	-> \fnm{Joergen W.}
%% Particle	-> \spfx{van der} -> surname prefix
%% FamilyName	-> \sur{Ploeg}
%% Suffix	-> \sfx{IV}
%% NatureName	-> \tanm{Poet Laureate} -> Title after name
%% Degrees	-> \dgr{MSc, PhD}
%% \author*[1,2]{\pfx{Dr} \fnm{Joergen W.} \spfx{van der} \sur{Ploeg} \sfx{IV} \tanm{Poet Laureate}
%%                 \dgr{MSc, PhD}}\email{iauthor@gmail.com}
%%=============================================================%%

\author[1]{\fnm{Shuai} \sur{Bian}}\email{2071207@stu.neu.edu.cn}

\author[1]{\fnm{Shouliang} \sur{Qi}}\email{qisl@bmie.neu.edu.cn}

\author[1]{\fnm{Chen} \sur{Li}}\email{lichen@bmie.neu.edu.cn}

\author[2]{\fnm{Yudong} \sur{Yao}}\email{Yu-Dong.Yao@stevens.edu}

\author*[1,3]{\fnm{Yueyang} \sur{Teng}}\email{tengyy@bmie.neu.edu.cn}

\affil[1]{\orgdiv{College of Medicine and Biological Information Engineering}, \orgname{Northeastern University}, \city{Shenyang} \postcode{110169}, \state{China}}

\affil[2]{\orgdiv{Department of Electrical and Computer Engineering}, \orgname{Stevens Institute of Technology}, \orgaddress{\street{Hoboken}, \city{NJ} \postcode{07030}, \state{USA}}}

\affil[3]{\orgdiv{Key Laboratory of Intelligent Computing in Medicine}, \orgname{Ministry of Education}, \city{Shenyang} \postcode{110169}, \state{China}}

%%==================================%%
%% sample for unstructured abstract %%
%%==================================%%

\abstract{Deep learning has been applied to compressive sensing (CS) of images successfully in recent years. However, existing network-based methods are often trained as the black box, in which the lack of prior knowledge is often the bottleneck for further performance improvement. To overcome this drawback, this paper proposes a novel CS method using non-local prior which combines the interpretability of the traditional optimization methods with the speed of network-based methods, called NL-CS Net. We unroll each phase from iteration of the augmented Lagrangian method solving non-local and sparse regularized optimization problem by a network. NL-CS Net is composed of the up-sampling module and the recovery module. In the up-sampling module, we use  learnable up-sampling matrix instead of a predefined one. In the recovery module, patch-wise non-local network is employed to capture long-range feature correspondences. Important parameters involved (e.g. sampling matrix, nonlinear transforms, shrinkage thresholds, step size, $etc.$) are learned end-to-end, rather than hand-crafted. Furthermore, to facilitate practical implementation, orthogonal and binary constraints on the sampling matrix are simultaneously adopted. Extensive experiments on natural images and magnetic resonance imaging (MRI) demonstrate that the proposed method outperforms the state-of-the-art methods while maintaining great interpretability and speed.}


%%\pacs[JEL Classification]{D8, H51}

%%\pacs[MSC Classification]{35A01, 65L10, 65L12, 65L20, 65L70}

\maketitle

\section{Introduction}\label{sec1}

 Compressed sensing (CS) theory has received a lot of attention in recent years. CS proves that when a signal is sparse in a certain domain, it can be recovered with high probability from much fewer measurements than the Nyquist sampling theorem \cite{ref1,ref2,ref3,ref4,ref5}. The potential reduction in measurements is attractive for diverse practical applications, including but not limited to magnetic resonance imaging (MRI) \cite{ref6}, radar imaging \cite{ref7} and sensor networks \cite{ref8}.

Over the past decades, a great deal of image CS reconstruction methods have been developed based on sparse representation model \cite{ref9}, which operates on the assumption that many images can be sparsely represented by a dictionary. The majority of those traditional methods use some structured sparsity as an image prior and then solve a sparsity-regularized optimization problem in an iterative fashion \cite{ref10,ref11}. Some elaborate structures were introduced into CS, like Gaussian scale mixtures model in wavelet domain \cite{ref12}. In addition, non-local self-similarity of image is also introduced to enhance the CS performance \cite{ref13,ref14,ref15}. For example, Metzler $et$ $al$. \cite{ref16} combined a Block Matching 3D (BM3D) denoiser into approximate message passing (AMP) framework to perform CS reconstruction. Zhang $et$ $al$. \cite{ref13} proposed the method combining sparse prior with non-local regularizers achieved well performance. Recently, some optimization-based methods have implemented adaptive sampling using alternating optimization techniques to jointly optimize the sampling matrix and CS recovery algorithms \cite{ref9}. Despite the excellent interpretability of the above methods, they all require hundreds of iterations to produce decent results, which inevitably entails a heavy computational burden, in addition to the challenges posed by hand-craft transformations and associated hyper-parameters.

Inspired by the successful applications of deep learning, several network-based CS reconstruction methods were developed to learn the inverse mapping from the CS measurement domain to original signal domain \cite{ref17,ref18,ref19}. Mousavi $et$ $al$. \cite{ref20} applied a stacked denoising auto-encoder (SDA) to learn the statistical relationship from training data. However, the fully connected network used in SDA results in high computation cost. Kulkarni $et$ $al$. \cite{ref21} developed a method based on convolutional neural networks, called Recon-Net, to reconstruct the original image from the CS sampled image blocks. Yao $et$ $al$. \cite{ref22} used residual learning to further improve CS reconstruction. Sun $et$ $al$. \cite{ref23} propose a novel sub-pixel convolutional generative adversarial network (GAN) to learn compressed sensing reconstruction of images. To mitigate block effect in reconstruction, some models make use of full image areas for reconstruction \cite{ref24,ref25}. Meanwhile, for further improving the CS performance, some models are proposed to train the non-linear recovery operator to learn the optimal sampling pattern with recovery model \cite{ref26,ref27,ref28}. The main advantage of the network-based methods is the reconstruction speed, as opposed to their optimization-based counterparts. However, the barrier to future performance improvement is their lack of the CS domain-specific insights intrinsic to optimization-based approaches.

To overcome above shortcomings, researchers link optimization methods to networks, which make them interpretable. Specifically, these methods embed the solving process of traditional optimization-based methods into the forward operator of deep learning. For instance, Zhang $et$ $al.$ \cite{ref29}
 proposed a deep network called ISTA-Net, which maps the popular Iterative Shrinkage Thresholding (ISTA) algorithm to network. It learns the sparse transform and soft threshold function via network. Based on ISTA-Net, Zhang $et$ $al$. \cite{ref30} proposes Opine-Net, which combines an efficient sampling module with ISTA-Net to achieve adaptive sampling and recovery. More recently, You $et$ $al$. \cite{ref31} improved ISTA-Net, which enables a single model to adapt for multiple sampling rates. Xiang $et$ $al$. \cite{ref32} proposed FISTA-Net for solving inverse problem, which is an accelerated version of ISTA. The Alternating Direction Method of Multipliers (ADMM) is proposed for the saddle point problem containing Lagrange multipliers that can not be solved directly by the ISTA algorithm. Drawing on the same idea, Yang $et$ $al$. \cite{ref33} proposed the ADMM-Net, which unfolds ADMM into network and applies it to CS-MRI. It employs a learnable transformation and the corresponding hyper-parameters in ADMM are learned from the network. Zhang $et$ $al$. \cite{ref34} extended the well-known AMP algorithm to propose AMP-Net. These models enjoy the interpretability with speed and tuning-free advantage. But the existing those approaches make little use of the non-local self-similarity pior which plays an important role in image reconstruction.
\begin{figure}
 \centering
\includegraphics[width=13cm,height=10cm]{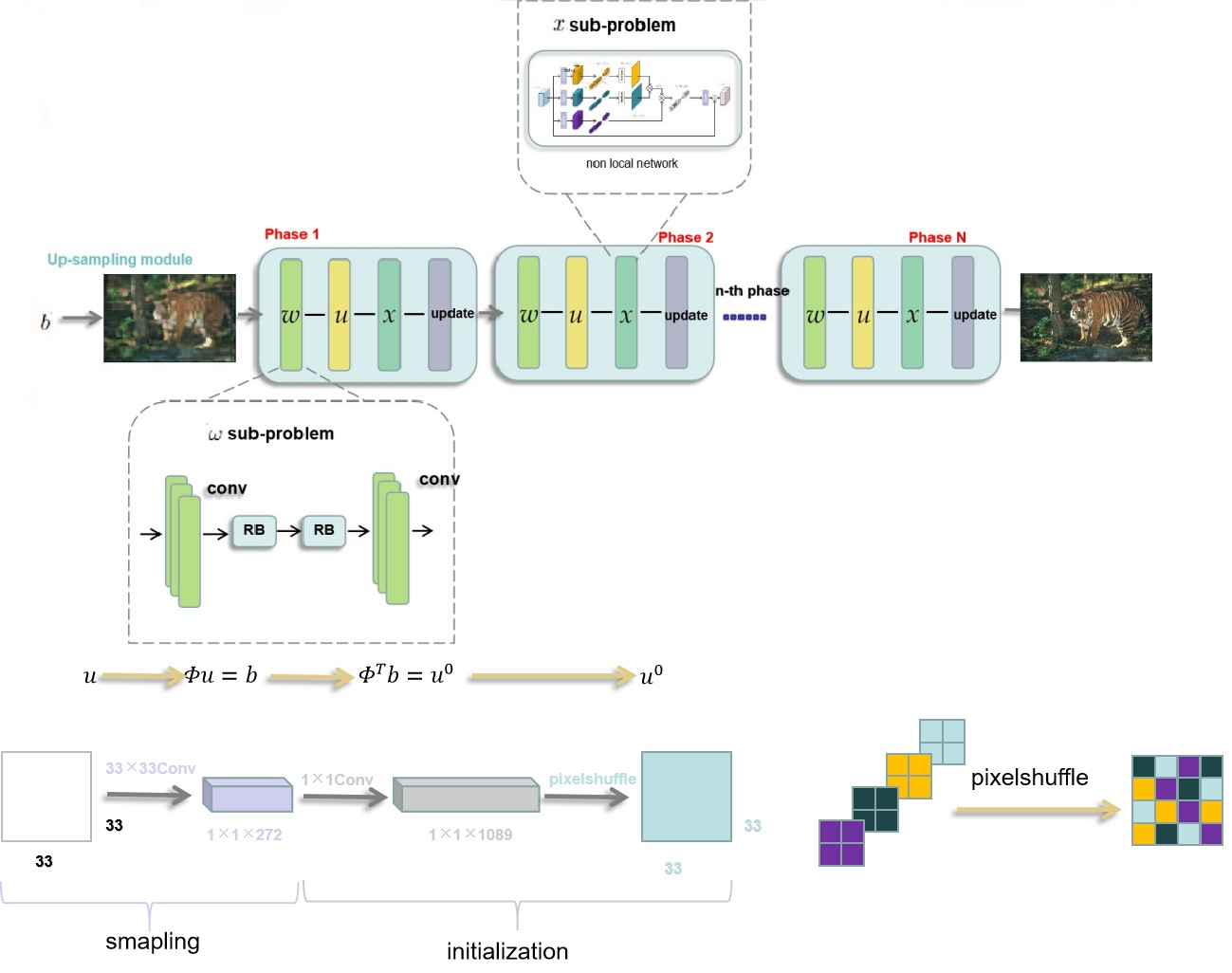}\\
\caption{ Illustration of recovery module in our proposed NL-CS Net. Specifically, NL-CS Net is composed of $N_p$ phases, and each phase corresponds to one iteration. $w$ module, $u$ module and $x$ module in $N_p$ phase corresponds to the solution of three sub-problems. The bottom half of the figure 1 is illustration of up-sampling module and the PixelShuffle operation.}\label{fig:1}
\end{figure}

There have been many previous methods for image reconstruction based on non-local prior. The non-local means (NLM) filter \cite{ref35} is highly successful in the image denoising, where it produces a denoised image by calculating the weighted value of the current pixel and its neighbouring pixels. Inspired by NLM, several inverse problem  frameworks incorporating non-local regularizer have been proposed \cite{ref13,ref14,ref15}. For instance, Zhang $et$ $al$. \cite{ref13} combines the TV regularizer with the non-local regularizer, which is solved using the augmented Lagrangian method and captures non-local features of the image during the iterative process. However, the use of time-consuming NLM filters in the iterations undoubtedly introduces a costly computational complexity. Inspired by deep learning, some recent network-based approaches exploit non-local self-similarity. Liu $et$ $al$. \cite{ref36} proposed a network incorporated non-local operations into a recurrent neural network (RNN) for image restoration. Although non-local prior has been widely exploited by both optimization-based and network-based methods, few interpretable deep learning models have introduced this important prior.

This paper combines merit of CS and non-local prior to propose a novel interpretable network, dubbed NL-CS Net. It composed of two parts: the up-sampling module and the recovery module. In the up-sampling phase, we adopted fully connection matrix to stimulate block-wise sampling and initial process. In the recovery phase, we maps the augmented Lagrangian method solving non-local regularized CS reconstruction model into the network, where the network consists of fixed number of phases, each of which corresponds to the one iteration. Rather than the traditional time-consuming NLM operation, the patch-wise non-local network is used to exploit global features. The hyper-parameters involved in NL-CS Net (e.g. sampling matrix, step size, $etc.$) are learned end-to-end, rather than being hand-crafted. Experimented results on natrual images dataset and MRI dataset shows the feasibility and effectiveness of the proposed method compared with the existing methods.
\section{Related work}\label{sec2}
 The goal of CS is to reconstruct image from its CS measurement with high quality. Mathematically, given the original image $u\in \mathbb R^N$, its CS measurements can be obtained by $b = \Phi u\in \mathbb R^M$, where $\Phi \in \mathbb R^{M\times N}$ denotes the sampling matrix and ${M}\diagup{N}$ (${M<<N}$) is commonly regarded as the CS sampling rate. Reconstructing $u$ from $b$ is typically ill-posed. Proposed NL-CS Net combines merit of CS and non-local prior, thus, we first review the traditional optimization-based algorithm to solve the non-local regularized model for CS.

The traditional methods use a preset sampling matrix to recover $u$ from the measured image $b$, which is formulated as solving  the following optimization problem:
\begin{equation}
\mathop {\min }\limits_u R(Du){\kern 1pt} {\kern 1pt} {\kern 1pt} {\kern 1pt} {\kern 1pt} s.t.{\kern 1pt} {\kern 1pt} {\kern 1pt} {\kern 1pt} {\kern 1pt} b = \Phi u\label{eq1}
\end{equation}
where $D$ denotes the transform matrix and $R$ is a regularizer, that imposes prior knowledge, such as sparsity and non-local self-similarity.

More effective in suppressing staircase artifact and restoring the detail, optimization-based approaches, combined traditional sparse priors with non-local regularizer, have been proven to achieve superior performance \cite{ref13}.
\begin{equation}
\mathop {\min }\limits_u {\left\| {Du} \right\|_1} + \alpha {\sum\limits_{{i}} {\left( {{u_i} - \sum\limits_{{j}} {{W_i}_j{u_j}} } \right)} ^2}s.t.\Phi u = b\label{eq2}
\end{equation}
The  ${W_i}_j$ is of the following form:
\begin{align}
\nonumber\\
{W_i}_j &= \left\{ \begin{array}{l}
{\exp (\left\| {{u_i^{(k)}} - {u_j^{(k)}}} \right\|_2^2/h)/c}{\kern 1pt} {\kern 1pt} {\kern 1pt} {\kern 1pt} {\kern 1pt} if{\kern 1pt} {\kern 1pt} {\kern 1pt} {j} \in {s_i}\\
0,{\kern 1pt} {\kern 1pt} {\kern 1pt} {\kern 1pt} {\kern 1pt} {\kern 1pt} {\kern 1pt} {\kern 1pt} {\kern 1pt} {\kern 1pt} {\kern 1pt} {\kern 1pt} {\kern 1pt} {\kern 1pt} {\kern 1pt} otherwise{\kern 1pt}
\end{array} \right.
\label{eq5}
\end{align}
where $s_i$ is the set containing the neighbor of the pixel $i$; ($W_{ij}$) represents the matrix form of NLM; $\alpha$ is a hyper-parameter; $h$ is a controlling factor; the superscript $k$ indicates the number of iterations. In brief, for a given pixel, the NLM filter can be obtined by calculating a weighted average of the surrounding pixels within a search window.

In order to solve Eq. (2), we equivalently transform Eq. (2) into the following problem through variable splitting technique.
\begin{equation}
\begin{array}{l}
\mathop {\min }\limits_{\omega ,u,x} {\left\| \omega  \right\|_1} + \alpha \left\| {x - Wx} \right\|_2^2\\
s.t.{\kern 1pt} {\kern 1pt} {\kern 1pt} \Phi u = b,u = x,Du = \omega
\end{array}\label{eq6}
\end{equation}
where $\omega $ and $x$ are auxiliary variables. Thus, the corresponding augmented Lagrangian function for Eq. (4) is expressed as:
\begin{align}
L(\omega ,u,x,v,\gamma,\lambda) =& {\left\| \omega  \right\|_1}+ \alpha \left\| {x - Wx} \right\|_2^2\nonumber\\
 &- {v^T}(Du - \omega )+\dfrac{\beta }{2}\left\| {Du - \omega } \right\|_2^2\nonumber \\
&- {\gamma ^T}(u - x) + \dfrac{\theta }{2}\left\| {u - x} \right\|_2^2\nonumber \\
&- {\lambda ^T}(\Phi u - b)+ \dfrac{\mu }{2}\left\| {\Phi u - b} \right\|_2^2
\end{align}
where $\theta$, $\mu$ and $\beta$ are regularization hyper-parameters; $\lambda$, $v$
 and $\gamma$ are the Lagrangian multipliers. In this case, the augmented Lagrangian method solves Eq. (5) by the folloing update rule:
\begin{equation}
({\omega ^{k + 1}},{u^{k + 1}},{x^{k + 1}}) = \mathop {\arg\min}\limits_{\omega,u,x~~~~} L(\omega ,u,x,v,\gamma,\lambda)\label{eq8}
\end{equation}
\begin{equation}
\left\{ \begin{array}{l}
{v^{(k + 1)}} = {v^{(k)}} - \beta (D{u^{(k + 1)}} - {\omega ^{(k + 1)}})\\
{\gamma ^{(k + 1)}} = {\gamma ^{(k)}} - \theta ({u^{(k + 1)}} - {x^{(k + 1)}})\\
{\lambda ^{(k + 1)}} = {\lambda ^{(k)}} - \mu (\Phi {u^{(k + 1)}} - b)
\end{array} \right.\label{eq9}
\end{equation}

    By applying the alternating direction method, Eq. (6) can be decomposed into three sub-problems in the following form:
\begin{align}
{\omega ^{(k + 1)}} =& S\left( {D{u^{(k)}} - \frac{{{v^{(k)}}}}{\beta },\frac{1}{\beta }} \right)\\
{u^{(k + 1)}} =& {u^{(k)}} - \varepsilon d\\
{x^{(k + 1)}} =& \frac{{\theta({u^{(k + 1)}} - \frac{\gamma^{(k)} }{\beta }) + 2\alpha W({u^{(k + 1)}} - \frac{\gamma^{(k)} }{\beta })}}{{\theta  + 2\alpha }} \label{eq13}
\end{align}
 where $d = {D^T}(\beta D{u^{\rm{(k)}}} - {v^{(k)}} - \beta {\omega ^{(k + 1)}})- {\gamma ^{(k)}} + \theta ({u^{(k)}} - {x^{(k)}})+ (\mu {\Phi ^T}(\Phi {u^{(k)}} - \Phi b) - {\lambda ^{(k)}})$.
 
 $S( \cdot )$ is a nonlinear shrinkage function with the hyper-parameter $1/{\beta}$, where $S\left( {y,z} \right)=sign(y)max(\lvert y\rvert-\frac{1}{z}, 0)$.
 
 And $\varepsilon$ is the step size of the gradient descent method. The overall algorithm flow is shown in Algorithm 1.

 This algorithm uses time-consuming NLM operations in each iteration. It typically requires hundreds of iterations to achieve a satisfactory result, which suffers from a large amount of computation. The transform $D$, sampling matrix and step size are pre-defined, which is very challenging to detain hyper-parameter.
\begin{table}[h]
\begin{center}
\begin{minipage}{274pt}
\begin{tabular}{@{}llll@{}}
\toprule
\quad \textbf{Algorithm 1} Non-local Regularized CS Algorithm
\\

\midrule

\quad \textbf{Input}:The sampled signal $b$ and sampling matrix $\Phi$ and $\alpha$, $\beta, \theta, \mu$ given.\\
\quad \textbf{Output}: $u$\\
\quad Initialization:$u^0=\Phi^T b $, $v^{(0) }=\gamma^{(0) } =\lambda^{(0) }=0$,  $x^0=\omega^0=0$
\\

\quad \textbf{While} (Outer stop conditions not satisfied) \textbf{do}\\
    \qquad \textbf{While} (Inner stop conditions not satisfied) \textbf{do}\\
       \quad\quad\quad Solve $\omega$ sub-problem by computing Eq. (8).\\
       \quad\quad\quad Solve $u$ sub-problem by computing Eq. (9).\\
        \quad\quad\quad Solve $x$ sub-problem by computing Eq. (10).\\
       \quad\quad\quad \textbf{End while}\\
       \quad\quad\quad upate multipliers $v, \gamma, \lambda$ by computing Eq. (7).\\
\quad\textbf{End while}\\
\botrule
\end{tabular}
\end{minipage}
\end{center}
\end{table}

\section{Proposed NL-CS Net for CS}\label{sec3}
 On the basis of Algorithm 1, by combining the merits of optimization-based and network-based approaches, the main idea of this paper is to unfold the solution process of  Algorithm 1 into network. The structure of NL-CS Net is shown in Figure 1. For jointly optimizing the sampling matrix and the recovery algorithm, our proposed NL-CS Net consists of a up-sampling module and a recovery module. In the up-sampling phase, we adopted a fully connection matrix to stimulate block-wise sampling and initial process. In the recovery phase, its backbone is designed by mapping the augmented Lagrangian method solving non-local regularized CS reconstruction model into network. The network consists of fixed number of phases, each of which corresponds to the one iteration. Hence, NL-CS Net is composed of $\omega^{(k)}$, $u^{(k) }$ and $x^{(k)} $ modules and update Lagrange multiplier module corresponding to four sub-problem Eqs. (8) , (9) , (10) and (7) sequentially in the $k$-th iteration. We designed a novel model to replace soft-shrinkage function and allowed step size and transform matrix to be learned. The learnable patch-wise non-local method is used to exploit global features, rather than the traditional non-local means operation. The parameters involved in NL-CS Net (e.g. sampling matrix, step size, $etc.$) are learned end-to-end, rather than being hand-crafted.

\subsection{Up-sampling module in NL-CS Net}\label{subsec5}
A warm start often leads to better result. The measured image $b$ is compressive from the original image $u$. We obviously use the sampling matrix ${\Phi }$ to obtain $b$ from $u$ as $b={\Phi }u$. Notice that, in this section, we indiscriminately use ${u,b,w,x}$ as one- or two-dimensional tensor according to actual demand. For example, in the formulation $b={\Phi }u$, $b$ and $u$ are one-dimensional, however, in a network, they are two-dimensional. Meanwhile, we can also obtain an approximate $u$ from $b$ as ${u}={\Phi^T }{{b}}$. In this module, ${\Phi }$ will be leanable instead of pre-defined.

It is well known that the linear transformation can be performed by a series of convolutional operators. Thus, we implement this operation by a convolutional layer and a PixelShuffle layer \cite{ref37}, specifically, adjusting the transpose of the sampling matrix $\Phi$ to $N$ filters with the same size $1\times1 \times {M}$. With those filters, ${{u}}={\Phi^T }{{b}}$ is implemented through a $1\times1 $ convolutional layer. PixelShuffle layer expands feature maps by reorganization between multiple channels, and we apply it to transform the tensor shape $1\times1 \times$ N output into $\sqrt{N}\times\sqrt{N} \times 1$.
\begin{equation}
{u^0}={\rm{ PixelShuffle }}\left( {{{\Phi ^T}}}{{b}} \right)\label{eq15}
\end{equation}
Obviously, $b$ is one-dimensional and $u^{0}$ is two-dimensional, and $u^{0}$ can be inputed into an image-targeted network.
\subsection{$\omega^{(k)}$, $u^{(k) }$ and $x^{(k)} $ module in NL-CS Net}\label{subsec5}
 In the following, we consider that the above three sub-problem Eqs. (8) , (9) and (10) in the $k$-th  iteration, and we unfold them into three separate modules in $k$-th phase of NL-CS Net: $\omega^{(k)}$ module, $u^{(k)}$ module and $x^{(k)}$ module.

The $\omega^{(k)}$ module corresponds to the Eq. (8) and is used to produce the output $\omega^{(k+1)}$. The transform matrix $D$ of traditional approach is to use a set of pre-trained filters. Here, we adopt a set of learnable filters to transform the image into the transform domain instead of the hand-crafted strategy. Note that it is hard to tune a well-designed threshold $1/{\beta}$ in Eq. (8) which is necessary to recover the details of the image. Hence, we set ${\beta}$ as learnable parameter. For efficiently solve the Eq. (8), we propose a flexible model for solve nonlinear transformation. In detail, the deep learning solution of the $\omega^{(k)}$ sub-problem can be described as follows:
\begin{equation}
{\omega ^{(k + 1)}} = F_2^{(k)}\left( {RB_2^{(k)}\left( {RB_1^{(k)}\left( {F_1^{(k)}\left( {E_1({u^{(k)}}) - \frac{{{v^{(k )}}}}{\beta }} \right)} \right)} \right)} \right)\label{eq16}
\end{equation}
 \begin{figure}
 \centering
\includegraphics[width=10.5cm]{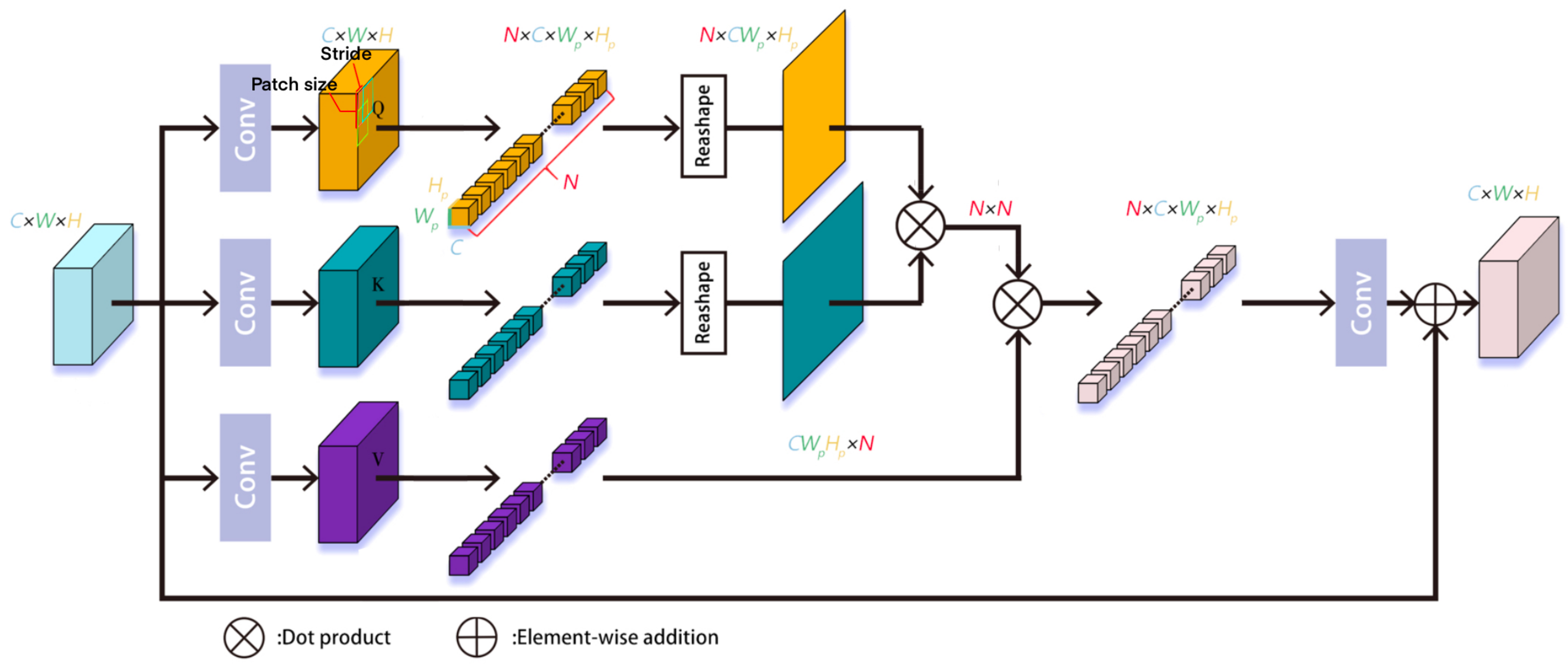}\\
\caption{  Illustration of patch-wise non-local network. We extract sliding local patches from input feature map.}\label{fig:3}
\end{figure}
Here, $E_1^{(k) }$ consists of size $3\times 3$ convolutional layer, which corresponds to 32 filters and Rectified Linear Unit (ReLU). To extract the features of the image and reconstruction, Eq. (12) is composed of two convolutional layers ($F_1^{(k)}$ and $F_2^{(k)}$) and two residual blocks ($RB_2^{(k)}$ and $RB_1^{(k)}$) . $F_1^{(k)}$ and $F_2^{(k)}$ denote size $3\times 3$ convolutional layer which corresponds to 32 filters and the residual blocks contain two $3\times 3$ convolutional layers which correspond to 32 filters and ReLU with skip connection from input to output.

Corresponding to gradient descent-based Eq. (9) in the $u^{(k) }$ module. We allow the step size to be learned in the network which is very different from the fixed step size of traditional methods. The $u^{(k) }$ module is  finally defined as:
\begin{align}
{u^{(k + 1)}} =& {u^{(k)}} - \varepsilon d\nonumber\\
d =& E_2^{(k)}(\beta E_1^{(k)}({u^k}) - {v^{(k)}} - \beta {\omega ^{(k + 1)}}) - {\gamma ^{(k)}} + \theta ({u^{(k)}} - {x^{(k)}}) \nonumber\\
&+ PixelShuffe[{\Phi ^T}(\mu (\Phi \mathop {{u^{(k)}}}   - \Phi \mathop {{u^0}}  ) - {\lambda ^{(k)}})]
\label{eq17}
\end{align}
where $E_2^{(k) }$composed of $3\times 3$ convolutional layer which corresponds to 32 filters and RELU.

We use the $x^{(k) }$ module to compute $x^{(k+1) }$ according to Eq. (10) with input $u^{(k+1) }$. For more efficient extraction of global features from images, the patch-wise non-local neural networks \cite{ref38} is used. It constructed the long-range dependence between image patches and applied a learnable embedding function to make the matching process adaptive. We use the learnable non-local method $NLM_{patch} ()$ instead of traditional NLM, as shown in Figure 2. In $NLM_{patch} ()$, given the input feature map $u^{(k+1) }$, we use three independently learnable weight matrices $F_Q$ ,$F_K$ and $F_V$ as the embedding functions which is implemented as $1\times 1$ convolution operation corresponds to 32 filters on the entire feature map. Instead of performing pixel-wise similarity computation in the embedded feature map directly like \cite{ref39}, a sliding window with a size of $7\times 7$ and a step size of 4 is used to select the overlapping patches in the embedded feature map. After the patch extraction operation, we have three sets of patches with size $N\times C\times W\times H$, so that our weight update strategy is to calculate the similarity between those patches. Next, we reshape the patch under the $F_Q$ and $F_K$ to a one-dimensional patch. $M$ denoted the temporary results, which can be calculated as follows:
\begin{equation}
{M} = {\mathop{\rm softmax}\nolimits} \left( {F_Q^T\left( {u^{(k + 1)}} - \frac{\gamma^{(k)} }{\beta } \right){F_K}\left( {u^{(k + 1)}} - \frac{\gamma^{(k)} }{\beta } \right)} \right).\label{eq18}
\end{equation}
 In the next step, we calculate the dot product of $F_V  (r)$ and $M$. Then, we recover these patches into the feature map of size $C\times W\times H$ with using averaging to process the overlapping areas. Finally, we place the output tensor through a convolutional layer and set up a skip connection between it and the input. Combining the $NLM_{patch} ({u^{(k + 1)}} - \frac{\gamma^{(k)} }{\beta } )$ with Eq.(10) yields the $x^{(k) }$ module as follows:
\begin{equation}
{x^{(k + 1)}} = \frac{{\theta ({u^{(k + 1)}} - \frac{\gamma^{(k)} }{\beta })+ 2\alpha N{{{\mathop{\rm LM}\nolimits} }_{patch}}\left( {u^{(k + 1)}} - \frac{\gamma^{(k)} }{\beta } \right)}}{{\theta  + 2\alpha }}.\label{eq19}
\end{equation}
Finally, we update the Lagrangian multiplier at each phase, which is the same with Eq. (7).

\subsection{Total loss function}\label{subsec10}
  We will show how to incorporate the two constraints with regarded to $\Phi$ into NL-CS Net simultaneously, including the orthogonality constraint and the binary constraint \cite{ref30}. For the orthogonal constraint $\Phi\Phi^T=I$, where $I$ is the identity matrix, the orthogonal loss term is defined as ${L_{orth}} = \frac{1}{{{M^2}}}\left\| {\Phi {\Phi ^T} - I} \right\|_F^2$, where $\left\| { \cdot } \right\|_F^2$ stands for the Frobenius norm, and we add this directly into the loss function.

To facilitate practical application, we restrict the value of the sampling matrix to 1 or 0. Binary($\cdot$) performs the following  operation on each element.
\begin{equation}
Binary(z)\left\{ {\begin{array}{*{20}{c}}
1&{if{\kern 1pt} {\kern 1pt} {\kern 1pt} {\kern 1pt} z \ge 0,}\\
0&{if{\kern 1pt} {\kern 1pt} {\kern 1pt} {\kern 1pt} {\kern 1pt} z < 0.}
\end{array}} \right.
\end{equation}
As previously described, we have successfully mapped the process of solving Eq. (2) to our NL-CS Net. The learnable parameters in NL-CS Net are defined in Table 1.

\begin{table}[h]
\begin{center}
\begin{minipage}{295pt}
%\textbf{Table 1} leanable parameters.\\ \label{tab1}%
\caption{leanable parameters.}\label{tab1}%
\begin{tabular}{@{}lllllllllllllll@{}}
\toprule
\multicolumn{15}{c}{Leanable parameters}\\
\midrule
$w$ modulee&&&~&~&&~&~& $RB_1$,$RB_2$,$F_1$ and $F_2$&~&~&&~&~&\\
$u$ module&&&~&~& &~&~&$E_1$,$E_2$, and $\varepsilon $&~&~&&~&~&\\
$x$ module&&&~&~&&~&~& $F_Q$,$F_k$ and $F_v$&~&~&&~&~&\\
Others &&&~&~&&~&~&$\Phi$,$\alpha$,$\beta$, and $\mu$&~&~&&~&~&\\
\botrule
\end{tabular}
\end{minipage}
\end{center}
\end{table}

Note that all those parameters are learned end-to-end rather than handcraft. Recovery module are not shared parameter across phase by default, which is a significant difference from traditional optimization-based algorithms.

Given a dataset $\{{u_1,u_2,u_3,...,u_{N_b }}\}$ where $N_b$ is the number of image blocks and $u_{i}$ represents the original image block, the output of the network through $N_p$ phase is denoted as $u_{i}^{(N_p ) }$. Our aim is to minimize the discrepancy between the network output $u_{i}^{(N_p ) }$ and the original image $u_i$ while satisfying the orthogonal constraint and the binary constraint. Hence the loss function of NL-CS Net is defined as follows:
\begin{align}
\min {\kern 1pt} {\kern 1pt} {\kern 1pt} {\kern 1pt} {\kern 1pt} {\kern 1pt} {\kern 1pt} {\kern 1pt} {L_{total}} &= {L_{{\rm{discrepency}}}}\; + \pi {L_{{\rm{orth}}}}\nonumber\\
s.t.~~~~Bina&ry(\Phi )\nonumber\\
where:{L_{{\rm{discrepency}}}} &= \dfrac{1}{{N{N_b}}}\sum\limits_{i = 1}^{{N_b}} {\left\| {u_i^{({N_p})} - {u_i}} \right\|} \nonumber\\
~~~~~~~~~~~{L_{{\rm{orth}}}} &=\dfrac{1}{{{M^2}}}\left\| {\Phi {\Phi ^T} - I} \right\|_F^2
\label{eq22}
\end{align}
 where $\pi$ is set to be 0.001 by experience.

\section{Experimental results}\label{sec4}
 We validate the proposed model on two tasks: the CS reconstruction of natrual images and MRI images. People are most often in contact with natural images, which is very important. MRI is a non-invasive and widely used imaging technique providing both functional and anatomical information for clinical diagnosis. But long scanning and waiting times may lead to motion artefacts and patient discomfort. MRI acceleration is one of the most successful applications of CS (CS-MRI), which can reconstructs high quality MR images from a few sampling data in k-space. To give quantitative criteria, Peak Signal to Noise Ratio (PSNR) is introduced to analyze the reconstruction performance.
 We use the Adam optimizer with the default learning rate set to 0.0001 and the batch size to 64. All the network were trained on a workstation configured with Intel Core i7-9700 CPU and RTX2080 GPU, and tested on a workstation configured with Intel Core i7-7820 CPU and GTX1060 GPU.
\subsection{Experiment on natural image}\label{subsec11}
 The training set was standardized using train90 \cite{ref21}, which contains 90 natural images. They are constructed by 88,912 randomly cropped image blocks (each of size $33\times33$). The corresponding measurement matrix is obtained from the training as opposed to fix it. The widely used benchmark datasets: Set11 \cite{ref21} and BSD68 \cite{ref40}, which have 11 and 68 natural images, respective, were applied for testing. The reconstruction results are presented as the average PSNR of the test images.
\subsubsection{ Hyper-parameter selection: phase and epoch numbers}\label{sec13}
To probe the appropriate number phase $N_p$ for NL-CS Net, we set the phase number $N_p$ from 1 to 15 to observe its performance, in the cases of 25\% CS sampling rate reconstruction on Set11. As can be seen in Figure 3a, PSNR rises gradually with the phase number. The curve is almost flat when $N_p \ge 10$.  To achieve a balance between performance and computational cost, in the following experiments, we set $N_p= 9$. Figure 3b further demonstrates the convergence process for three types of losses (i.e $L_{discrepency}$, $L_{orth}$ and $L_{total}$). We experimented with $N_p =9$ at the sampling rate of 25\% on Set11. The orthogonal constraint term gradually converges to zero which proves its suitability for NL-CS Net. Total loss achieves an acceptable result at about 120 epochs and converges at about 200 epochs. As below, we set epoch number to be 200 for an enough convergence.
\begin{figure}
 \centering
\includegraphics[width=11.5cm]{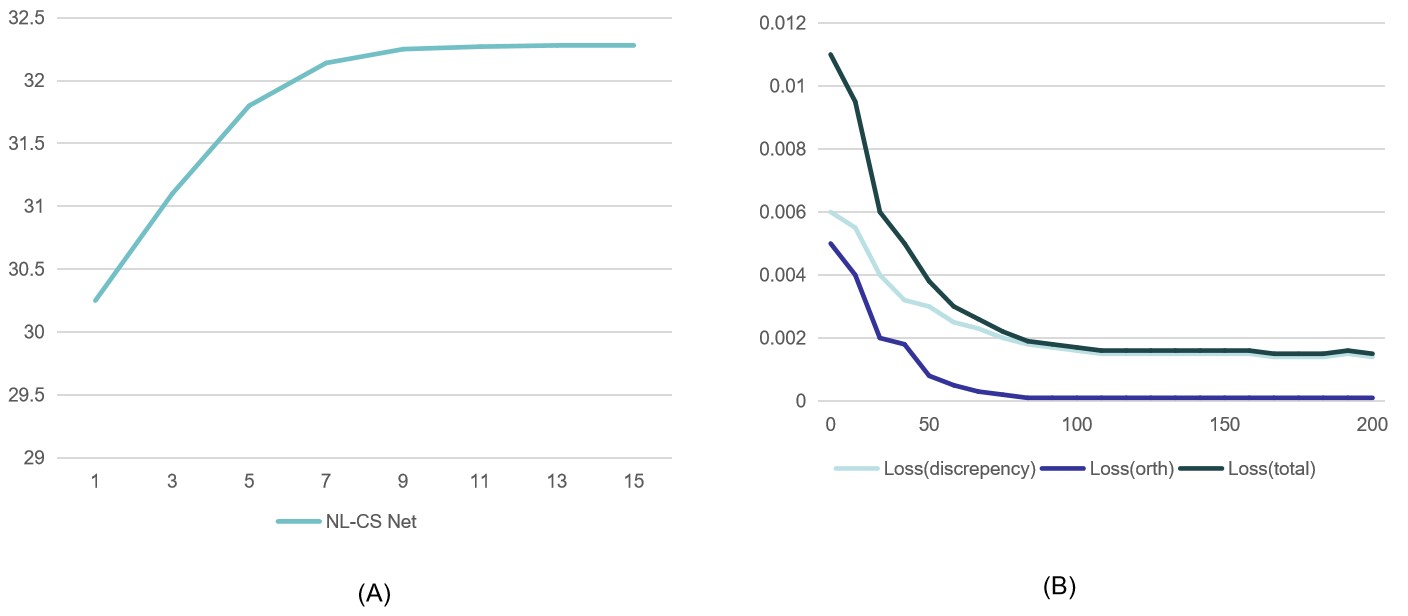}\\
\caption{ Phase number of NL-CS Net in the case of CS sampling rate = 25\% in 3A. The 3B progression curves of loss (discrepency) and loss (orth) achieved by NL-CS Net in training with various epoch numbers in the case of CS sampling rate = 25\% on Set11.}\label{fig:5}
\end{figure}
\subsubsection{Ablation studies}\label{subsec14}
\begin{table}[h]
\begin{center}
\begin{minipage}{284pt}
\caption{Ablation studies. The best performance is in bold.}\label{tab2}%
\begin{tabular}{@{}lllllll@{}}
\toprule
&\multicolumn{6}{c}{ \textbf { CS sampling rate (BSD68) } }\\
\textbf { Algorithm } &&&&&&\\
& 50 \% & 25 \% & 10 \% & 4 \% & 1 \% & \text { Avg }\\
\midrule
ISTA-Net & 34.04 &  29.36  &  25.32 &  22.17  &  19.14 & 26.01  \\
NL-CS Net(fixed $\Phi $ ) & 34.01 & 29.80 & 25.87 & 22.53&19.86&26.41\\
\textbf{NL-CS Net}&\textbf{34.69} & \textbf{29.97} & \textbf{26.72} & \textbf{24.21} & \textbf{21.63} & \textbf{27.44}  \\
\botrule
\end{tabular}
\end{minipage}
\end{center}
\end{table}

To adequately demonstrate the advantage of combining non-local regularized terms, we designed ablation experiments. ISTA-NET provides a network form soultion for the $L_1$ norm regularized optimization problem without the non-local regularized terms. For a fair comparison, we trained NL-CS Net with the Gaussian random sampling matrix as ISTA-NET, using the same training set, and tested its performance on BSD68 and the CS sampling rate varies in $\{1\%, 4\%,10\%,25\%,50\%\}$. As expected from Table 2, NL-CS Net with both fixed and varible sampling matrix outperforms ISTA-net, which further demonstrates the reasonableness of our method. In addition, we observe that joint optimized sampling matrix and recovery operator in our method improves performance by 1.4 over the fixed sampling matrix.

\begin{table}[h]
\begin{center}
\begin{minipage}{305pt}
\caption{The effect of different combinations on the reconstruction result.}\label{tab2}%
\begin{tabular}{@{}llllllll@{}}
\toprule
&\multicolumn{7}{c}{Different combinations of constraints of NL-CS Net }\\
\midrule
\text { Binary constraint }&  \checkmark && X && X && \checkmark \\
 \text { Orthogonality constraint } & X && \checkmark & &X && \checkmark \\
\textbf { PSNR } & \textbf {29.92} && \textbf {29.95} && \textbf {29.85} && \textbf {29.97} \\
\botrule
\end{tabular}
\end{minipage}
\end{center}
\end{table}

 NL-CS Net introduces two types of constraints including orthogonality constraint and binary constraint. We observe the effect of these two constraints on the reconstruction performance in the case of CS sampling rate = 25\%  on BSD68. It can be seen in Table 3 that orthogonality constraint and binary constraint acts as network regularization, which enhance the reconstruction performance.

In \begin{figure}
 \centering
\includegraphics[width=9.5cm]{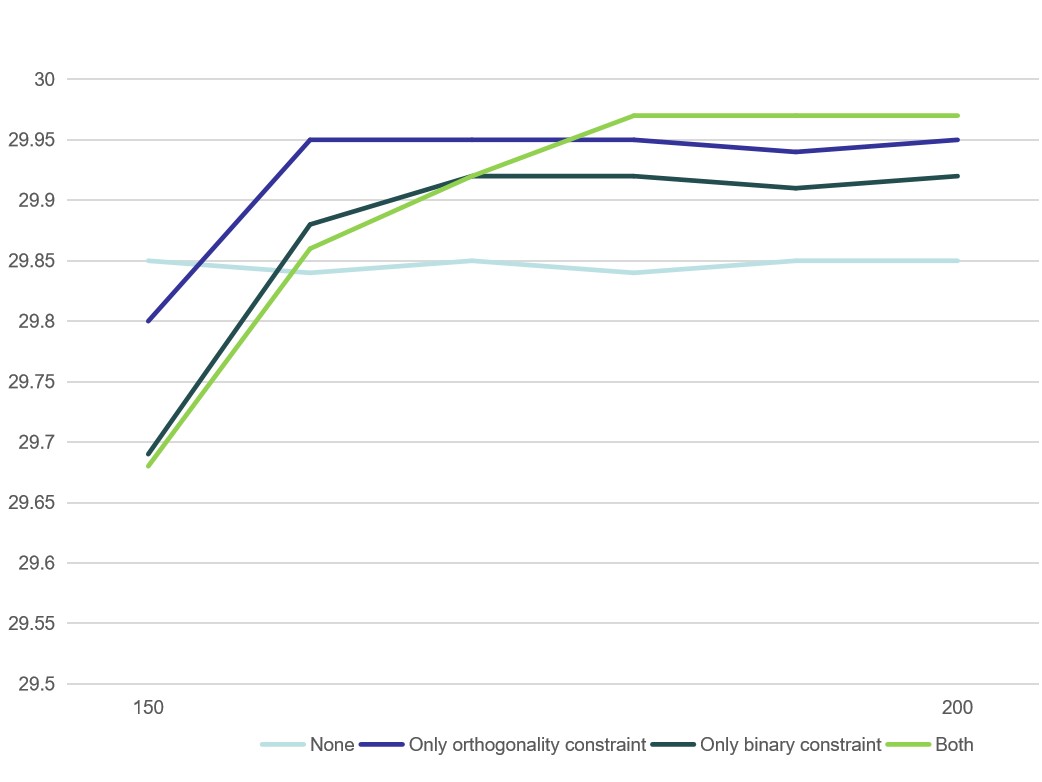}\\
\caption{   The curves for each combination are based on the PSNR in the case of CS sampling rate = 25\%.}\label{fig:7}
\end{figure}
Figure 4, we verify the effect of different constrain combinations on the convergence process, and it can be observed that all combinations converge to similar values at 200 epochs and the combination of orthogonality constraint and binary constraint achieves the best results.
\begin{table}[h]
\begin{center}
\begin{minipage}{314pt}
\caption{PSNR performance comparisons on Set11 with different CS sampling rates. The best performance is in bold.
  Note that the last column is a run-time analysis of all the competing methods.}\label{tab2}%
\begin{tabular}{@{}llllllll@{}}
\toprule
&\multicolumn{6}{c}{ \textbf { CS sampling rate (Set11) } }&\textbf{Time}\\
\textbf { Algorithm } &&&&&&&Cpu/GPU\\
& 50 \% & 25 \% & 10 \% & 4 \% & 1 \% & \text { Avg } &\\
\midrule
\text { TVAL3 } & 33.56 & 27.92 & 23.00 & 18.75 & 16.43 & 23.93 & 3.150 s/$--$ \\
 \text { D-AMP } & 35.93 & 28.47 & 22.64 & 18.40 & 5.20 & 22.13 & 51.21 s/$--$ \\
 \text { IR-CNN } & 36.23 & 30.07 & 24.02 & 17.56 & 7.78 & 23.13 & $--$/ 68.42 s \\
 \text { SDA } & 28.95 & 25.34 & 22.65 & 20.12 & 17.29 & 22.87 & $--$/ 0.003 s \\
 \text { ReconNet } & 31.50 & 25.60 & 24.28 & 20.63 & 17.29 & 23.86 & $--$/ 0.016 s \\
 \text { ISTA-Net } & 37.74 & 31.53 & 25.80 & 21.23 & 17.30 & 26.72 & $--$/ 0.093 s \\
 \text { FISTA-Net } & \textbf{37.85} & 31.66 & 25.98 & 21.20 & 17.34 & 26.81 & $--$/ 0.052 s \\
 \text { BCS } & 34.61 & 29.98 & 26.04 & 23.19 &  19.15 & 26.59 & $--$/ 0.002 s \\
 \text { NL-CS Net } & 37.29 & \textbf{32.25} & \textbf{27.53} & \textbf{24.04} & \textbf{19.60} & \textbf{28.13} & $--$/ 0.326 s \\
\botrule
\end{tabular}
\end{minipage}
\end{center}
\end{table}
\subsubsection{Comparison with state-of-the-art methods}\label{subsec11}
We compare the proposed NL-CS Net with eight representative models, including TVAL3 \cite{ref41}, D-AMP \cite{ref16}, IR-CNN \cite{ref42}, SDA \cite{ref20}, Recon-Net \cite{ref21}, ISTA-Net \cite{ref29}, FISTA-Net \cite{ref32} and BCS \cite{ref26}. TVAL3, D-AMP and IR-CNN are the optimization-based methods; Recon-net, SDA and BCS are the network-based methods. ISTA-Net and FISTA-Net are interpretable Network. In particular, IR-CNN inserts the trained CNN denoiser into the Half Quadratic Splitting (HQS) optimization method, which is used to solve the inverse problem. Recon-Net use convolutional method to learning the inverse problem map
and reconstruction. BCS learns the sampling matrix through the network. ISTA-Net and FISTA-Net are constructed by unfolding traditional optimization-based algorithms into deep learning.
\begin{figure}
 \centering
\includegraphics[width=11.5cm]{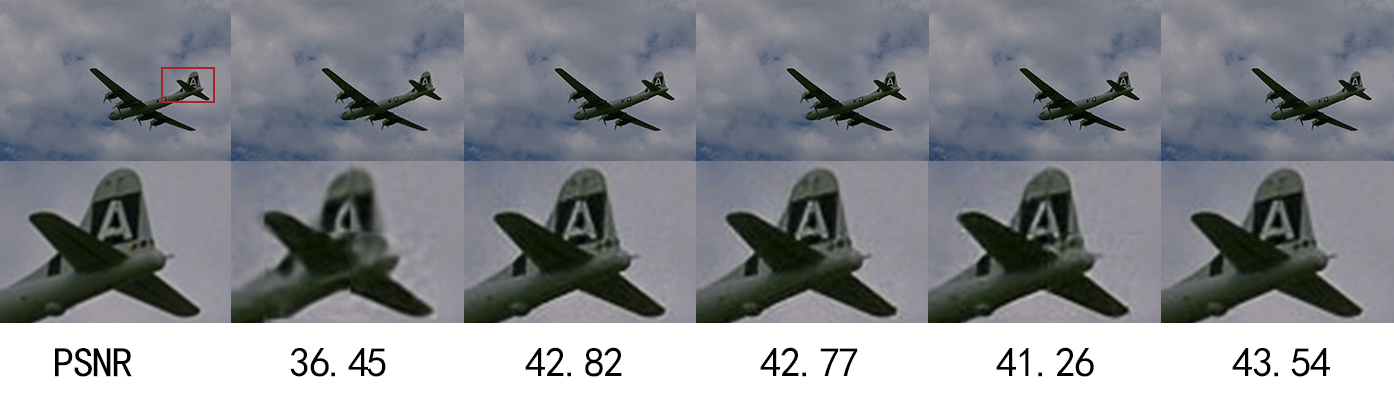}\\
\caption{ Average PSNR (dB) performance comparisons on BSD68. From left to right: original image, Recon-Net, ISTA-Net, FISTA-Net, BCS and NL-CS Net (ours).}\label{fig:3}
\end{figure}
Table 4 shows the quantitative results of various CS algorithms on Set11. For the optimization-based methods including TVAL3, D-AMP and IR-CNN, we observe that they perform badly at extremely low CS sampling rates of 1\%-4\%, which has a large gap in performance with the other two categories of algorithms. Meanwhile, the proposed NL-CS Net outperforms the optimization-based methods at all the sampling rates. Specifically, NL-CS Net achieves on average 4.2 gain against the best-performing optimization-based method (TVAL3). In particular, the proposed NL-CS Net achieves a gain of 3.17, 11.82, 14.4, over TVAL3, D-AMP and IR-CNN respectively at extremely low 1\% sampling rate. The network-based methods, including Recon-net, SDA and BCS, perform well at all sampling rates compared with the traditional methods. Still, at most sampling rates, NL-CS Net achieved the best results except that ISTA-Net and FISTA-Net obtain a minor advantage only with 50\% of CS sampling rate. Compared to the two state-of-the-art interpretable networks ISTA-Net and FISTA-Net, the proposed NL-CS Net obtained a gain of 1.32 and 1.35 respectively on the average. In addition, compared to the optimization-based approaches, the proposed NL-CS Net substantially reduces the computation time. The reconstruction speed is approximately more than 10 times faster than that of D-AMP and IR-CNN. Compared to network-based approaches, NL-CS Net achieves decent speed with best performance.

\begin{table}[h]
\begin{center}
\begin{minipage}{234pt}
\caption{PSNR (dB) performance comparisons on BSD68 with different CS sampling rates. Best performance is in bold.}\label{tab2}%
\begin{tabular}{@{}lllllll@{}}
\toprule
&\multicolumn{6}{c}{ \textbf { CS sampling rate (BSD68) } }\\
\textbf { Algorithm } &&&&&&\\
& 50 \% & 25 \% & 10 \% & 4 \% & 1 \% & \text { Avg }\\
\midrule
 \text { ISTA-Net } & {34.04}& {29.36}&25.32&22.17&19.14&26.01 \\
 \text { FISTA-Net } & 34.28&29.45&25.38&22.31&19.35&26.16 \\
 \text { BCS } &33.18&29.18&26.07&23.94&21.24& {26.72}\\
 \textbf { NL-CS Net } &   \textbf {34.69}&\textbf{29.97}&\textbf{26.72}&\textbf{24.21}&\textbf{21.63}&\textbf{27.44} \\
\botrule
\end{tabular}
\end{minipage}
\end{center}
\end{table}
To further validate the generalizability of our NL-CS Net, we experimented several models that performed well on Set11, including ISTA-Net, FISTA-Net, BCS and ours on a larger dataset BSD68. In Table 5, it can be clearly observed that NL-CS Net outperforms the other algorithms at all sampling rates. It outperforms the second best algorithm by 0.72 in average PSNR, and by 0.39, 0.27, 0.52, 0.61 and 0.41 for different sampling rates from 1\% to 50\%, respectively.

Figure 5 shows a visual comparison. As can be seen, NL-CS Net is capable of preserving more texture information and recovering richer structural detail due to the effective incorporation of the non-local prior.

\begin{figure}
 \centering
\includegraphics[width=11.5cm]{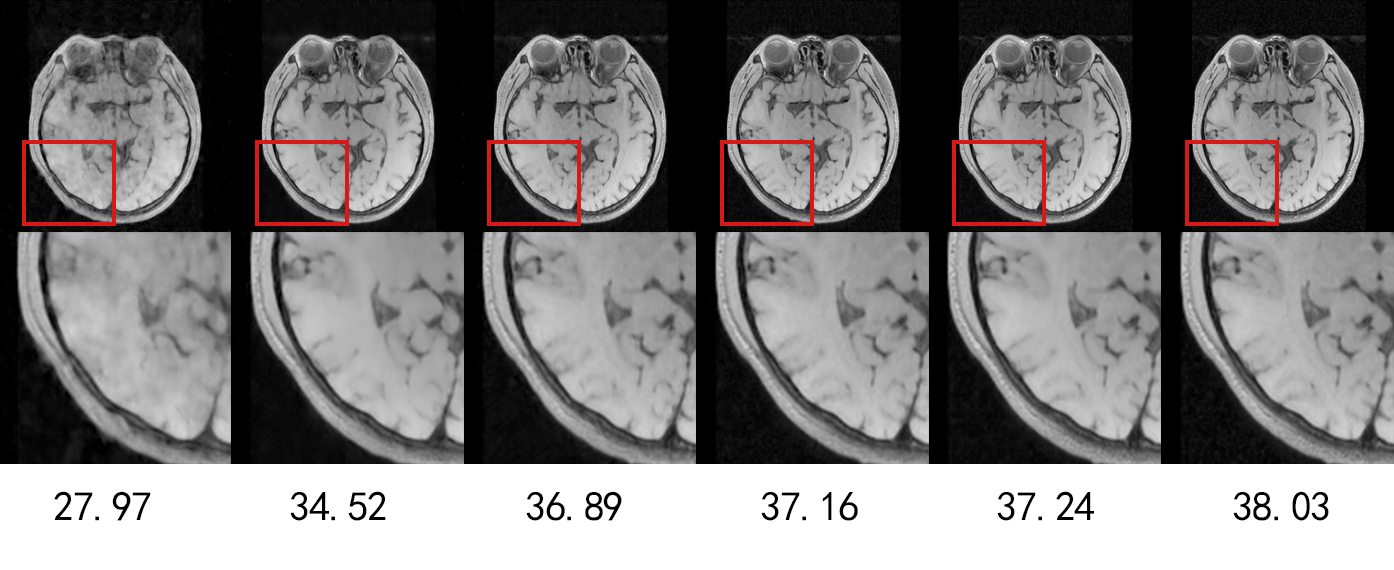}\\
\caption{   MRI reconstruction. From left to right: Zero-filling, RecPF , TV , PBDW ,U-Net and NL-CS Net.}\label{fig:6}
\end{figure}
\begin{table}[h]
\begin{center}
\begin{minipage}{254pt}
\caption{Average PSNR on MRI reconstruction. Best performance in bold.}\label{tab2}%
\begin{tabular}{@{}lllllll@{}}
\toprule
&\multicolumn{6}{c}{ \textbf { MRI} }\\
\textbf { Algorithm } &&&&&&\\
& 50 \% & 40 \% & 30 \% & 20 \% & 10 \% \\
\midrule
\text { Zero-filling } & 36.73 & 34.76 & 32.59 & 29.96 & 26.35 \\
\text { TV } & 41.69 & 40.00 & 37.99 & 35.20 & 30.90 \\
\text { RecPF } & 41.71 & 40.03 & 38.06 & 35.32 & 30.99 \\
 \text { PBDW } & 41.81 & 40.21 & 38.60 & 36.08 & 31.45 \\
 \text { UNet } & 42.20 & 40.29 & 37.53 & 35.25 & 31.86 \\
\textbf { NL-CS Net } & \textbf{42.38} & \textbf{40.32}  & \textbf{38.63} & \textbf{36.12} & \textbf{32.09} \\
\botrule
\end{tabular}
\end{minipage}
\end{center}
\end{table}
\subsection{CS-MRI}\label{sec4}
 We train and test on the brain and chest MRI images \cite{ref33}, in which the size of images is $256\times256$. For each dataset, we randomly take 100 images for training and 50 images for testing. In our experiments, we take $\Phi=fZ$, where $f$ is Fourier transform and $Z$ is down sampling matrix. Our proposed NL-CS Net can be directly applied to CS-MRI reconstruction. Here we compare NL-CS Net with four classical CS-MRI methods: Zero-filling, TV \cite{ref6}, RecPF \cite{ref43}, PBDW \cite{ref44} and UNet \cite{ref45} .

 It can be clearly observed in Table 6 that NL-CS Net outperforms the other algorithms at all sampling rates. It outperforms the second best algorithm by 0.64, 0.04, 0.03, 0.11 and 0.57 for different sampling rates from 10\% to 50\%, respectively. The visualization results are shown in Figure 6, it can be seen that NL-CS Net reconstructs the brain image better than other methods. More details of the brain texture are preserved and the edges are more clearly.

\section{Conclusion}\label{sec5}
Inspired by traditional optimization, we proposed a novel CS framework, dubbed NL-CS Net, with the incorporated learnable sampling matrix and non-local piror. The proposed NL-CS Net possesses well-defined interpretability, and make full use of the merits of both optimization-based and network-based CS methods.
Extensive experiments show that NL-CS Net have state-of-art performance while  maintaining great interpretability. For future work, one direction is to extend our proposed model to other image inverse problems, such as deconvolution and inpainting. Another one is to combine other iterative algorithms with deep learning.
\section*{Acknowledgments}
This work was supported by the Natural Science Foundation of Liaoning Province (2022-MS-114).

\section*{Declarations}

\begin{itemize}
\item Funding: Natural Science Foundation of Liaoning Province (2022-MS-114)
\item Conflict of interest/Competing interests: The authors declare that they have no known
competing financial interests or personal relationships that could have
appeared to influence the work reported in this paper.
\item Availability of data and materials: We hereby declare that all data and materials used in this study are publicly available with no restrictions. The data used in this research has been made publicly available and can be accessed directly via https://github.com/bianshuai001/NL-CS-Net. 
\item Code availability: We declare that the code used in this study is open-source and publicly available for unrestricted use. The code used in this research can be accessed via the link https://github.com/bianshuai001/NL-CS-Net. Anyone can retrieve, download, and use the code for non-commercial purposes, subject to appropriate attribution of the source.

\end{itemize}

%
%\bibliographystyle{IEEEtran}
%\bibliography{sn-bibliography}
\end{document}